\documentclass[conference]{IEEEtran}
\IEEEoverridecommandlockouts
\usepackage[letterpaper, left=.625in, right=.625in, top=.7in, bottom=.9in]{geometry}
\usepackage{cite}
\usepackage{amsmath,amssymb,amsfonts}
\usepackage{algorithmic}
\usepackage{graphicx}
\usepackage{textcomp}
\usepackage{xcolor}
\usepackage{romannum}
\usepackage{amsthm}
\usepackage{algorithmic}
\usepackage[ruled,vlined,linesnumbered]{algorithm2e}
\usepackage{float}
\usepackage{subfigure}
\usepackage{enumitem}
\SetAlFnt{\footnotesize}
\def\BibTeX{{\rm B\kern-.05em{\sc i\kern-.025emF b}\kern-.08em
T\kern-.1667em\lower.7ex\hbox{E}\kern-.125emX}}

\newcommand{\GreedyTensile}{{\it Greedy-Tensile }}
\newcommand{\GreedyLax}{{\it Greedy-Lax }}
\newcommand{\GreedySpectral}{{\it Greedy-Spectral }}

\begin{document}

\title{\huge Bridging Local and Global Knowledge: Cascaded Mixture-of-Experts Learning for Near-Shortest Path Routing}

\author{\IEEEauthorblockN{Yung-Fu Chen}
\IEEEauthorblockA{
\textit{The Ohio State University}\\
Columbus, OH, USA \\
chen.6655@osu.edu}
\and
\IEEEauthorblockN{Anish Arora}
\IEEEauthorblockA{
\textit{The Ohio State University}\\
Columbus, OH, USA \\
anish@cse.ohio-state.edu}
} 
\maketitle

\begin{abstract}
While deep learning models that leverage local features have demonstrated significant potential for near-optimal routing in dense Euclidean graphs, they struggle to generalize well in sparse networks where topological irregularities require broader structural awareness. To address this limitation, we train a Cascaded Mixture of Experts (Ca-MoE) to solve the all-pairs near-shortest path (APNSP) routing problem. Our Ca-MoE is a modular two-tier architecture that supports the decision-making for forwarder selection with lower-tier experts relying on local features and upper-tier experts relying on global features.  It performs adaptive inference wherein the upper-tier experts are triggered only when the lower-tier ones do not suffice to achieve adequate decision quality.  Computational efficiency is thus achieved by escalating model capacity only when necessitated by topological complexity, and parameter redundancy is avoided.
Furthermore, we incorporate an online meta-learning strategy that facilitates independent expert fine-tuning and utilizes a stability-focused update mechanism to prevent catastrophic forgetting as new graph environments are encountered. Experimental evaluations demonstrate that Ca-MoE routing improves accuracy by up to 29.1\% in sparse networks compared to single-expert baselines and maintains performance within 1\%--6\% of the theoretical upper bound across diverse graph densities.
\end{abstract}


\section{Introduction}

In graph routing ---especially when formulated for mobile, ad hoc and wireless networks--- achieving both scalability and accuracy fundamentally involves a trade-off between local and global knowledge. Machine learning models for graph routing typically choose one over the other, yielding distinct limitations. Models leveraging local features ---such as inter-node distance, as in \GreedyTensile \cite{chen2025knowledge}--- are computationally efficient and scalable because they require only locally available or immediate neighbor information. Although these models effectively predict next-hop forwarders in dense Euclidean graphs\footnote{Euclidean graphs serve as an effective model for mapping the physical topologies found in mobile, ad hoc, and wireless networks \cite{iyer2006topological}.}, outperforming standard geographic routing algorithms in that context, they struggle to generalize to sparse networks (density $\rho \! \leq \! 2$), where topological irregularities necessitate broader structural awareness \cite{ashvinkumar2022local}. Specifically, they lack the context necessary to detour around structural holes. Conversely, approaches that use global features ---such as spectral Laplacian-based features \cite{maskey2022generalized, he2023sheaf}--- can navigate these complex topologies, but they are often impractical for real-time distributed routing because of the prohibitive cost of maintaining the entire network state. Thus, a static reliance on local only or global only features fails to simultaneously satisfy the demands of efficiency and generalization.

To reconcile this trade-off, a Mixture of Experts (MoE) architecture \cite{mu2025comprehensive} offers a robust solution. By decomposing the routing policy into specialized modules, the framework circumvents the limitations inherent in monolithic models. Instead of requiring a single model to generalize across heterogeneous topologies, MoE enables ``selective inference": it prioritizes using one or more lightweight expert models for deciding the forwarder in routes where local knowledge suffices, and prioritizes activating one or more resource-intensive experts ---augmented with global context--- for other routes. This selection of experts serves to bridge the gap between local and global knowledge, preserving the low latency of local heuristics while still being able to handle complex topologies.

This approach draws inspiration from the significant success of MoE architectures in enhancing specialization and computational efficiency. Operating on the ``divide-and-conquer'' principle \cite{jacobs1991adaptive, jordan1994hierarchical}, MoE dynamically selects and engages only a subset of experts based on input characteristics. Recent architectures, such as DeepSeek \cite{liu2024deepseek}, have further refined this by emphasizing the decoupling of common knowledge from domain-specific nuances, facilitating a modular design where universal patterns are structurally isolated from complex, task-specific information. 

\begin{figure*}[thb!]
    \centering
     \subfigure
     {
        \includegraphics[width=0.95\textwidth]{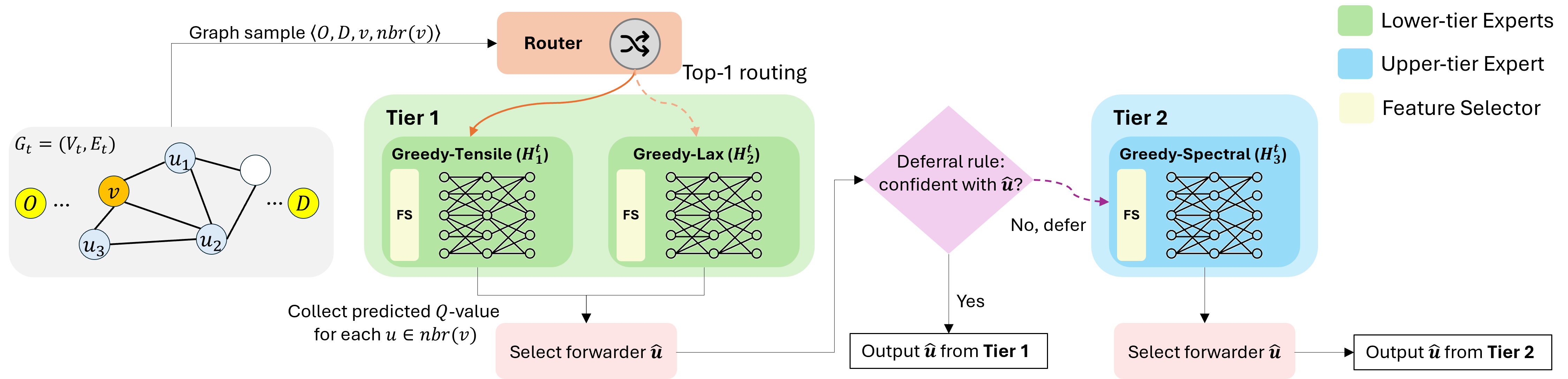}
    }
    \caption{The Ca-MoE framework for forwarder selection. The {\bf Router} directs the graph sample $\langle O,D,v,nbr(v) \rangle$ to a selected lower-tier expert that predicts a $Q$-value for each neighbor node $u$ in $nbr(v)$, where $nbr(v)$ denotes the set of $v$'s neighbors.  The candidate forwarder $\hat{u}$ for a near-shortest path from node $v$ to destination node $D$ is identified as a node in $nbr(v)$ with the highest $Q$-value. This candidate is validated by the {\bf Deferral rule}: the sample is cascaded to the upper-tier expert only if the confidence is insufficient. Dashed lines denote conditional execution paths, illustrating selective activation based on the graph sample (in orange) and the adaptive activation based on the candidate forwarder for the graph sample (in purple).}
    \label{Ca-MoE}
\end{figure*}


Moreover, modern MoE frameworks have increasingly adopted ``adaptive inference'' to optimize the trade-off between accuracy and efficiency \cite{kolawole2024agreement, nie2024online}. Unlike static activation, this paradigm employs a cascading strategy that initiates inference with a lightweight model tailored for common scenarios. A deferral rule\footnote{The deferral mechanism is typically implemented via a small neural network, which is computationally efficient, introducing negligible latency relative to the primary model execution.} is subsequently applied to assess prediction adequacy, triggering escalation to a resource-intensive expert only when necessary. This ensures substantial computational resources are allocated exclusively to complex scenarios.  We find that adaptive inference with such a Cascaded Mixture of Experts (Ca-MoE) architecture suffices to solve the all-pairs near-shortest path (APNSP) routing problem.

Although adaptive inference has proven effective in optimizing efficiency for computer vision \cite{kolawole2024agreement, kolawole2024revisiting} and language modeling \cite{raposo2024mixture}, its application to graph routing has not yet been explored. To the best of our knowledge, no existing research leverages machine learning to address the near-shortest path problem by learning an adaptive switching policy between local and global knowledge. Specifically, prior work has not established a mechanism that autonomously decides based on local knowledge only when to rely on inexpensive local heuristics versus when to escalate to costly global computations, which we demonstrate is feasible with our Ca-MoE solution.


To facilitate rapid adaptation to new graph topologies and reduce reliance on extensive pre-training, we employ an online meta-learning strategy that continuously updates both common and specialized experts as new graphs are encountered. To ensure robustness, we maintain an evaluation dataset that verifies whether fine-tuned experts retain performance on previously seen environments, thereby preventing catastrophic forgetting. Note that not all graphs are equally effective training seeds, this incremental learning process allows Ca-MoE to progressively accumulate diverse structural knowledge. Consequently, as the system observes more graphs, it converges toward higher accuracy and stronger generalization across a broad spectrum of graph classes.

The main contributions and findings of this paper are summarized as follows:
\begin{enumerate}[label=(\alph*)]
\item {\bf Cascaded Mixture of Experts (Ca-MoE) for solving graph routing:}
We adopt a modular multi-tier architecture that bridges the gap between local and global knowledge for the all-pairs near-shortest path (APNSP) problem. By utilizing adaptive inference between lightweight experts and conditionally activated resource-intensive experts, the framework optimizes both accuracy and computational efficiency.

\item {\bf Efficient online learning and continual adaptation:} We introduce an incremental learning strategy that fine-tunes experts independently using small sample sets. This approach facilitates rapid adaptation to new graph topologies and utilizes a stability-focused update mechanism (via an evaluation dataset) to prevent catastrophic forgetting as the network environment evolves.

\item {\bf Significant Performance Gains in Sparse Networks:} Experimental evaluations demonstrate that Ca-MoE significantly resolves the generalization limitations of single-expert baselines in sparse networks. Specifically, the framework improves routing accuracy by up to 29.1\% compared to local-feature models (e.g., {\em Greedy-Tensile}) by effectively navigating topological holes through specialized expert activation.

\item {\bf Consistency and Near-Optimality:} The framework maintains robust performance across diverse graph densities. It achieves up to a 5\% improvement in dense graphs and consistently maintains routing accuracy within 1\%–6\% of the theoretical upper bound (MoE\_OPT), demonstrating its ability to approximate optimal routing policies without incurring significant overhead.

\item {\bf Robustness to Initialization:} Our evaluations show that the online meta-learning strategy effectively mitigates the need for high-quality initial pre-training. Arbitrarily initialized experts are shown to rapidly converge to performance levels comparable to carefully optimized baselines after observing only a limited number of graphs.
\end{enumerate}


\section{Related Work}

\subsection{Machine Learning for Graph Routing}
Recent advances in knowledge-guided machine learning have demonstrated that deep neural networks (DNNs) can effectively solve the all-pairs near-shortest path (APNSP) problem by learning local routing policies \cite{chen2025knowledge}. Approaches in this domain generally fall into two categories: geometric and structural. Geometric methods, such as \GreedyTensile \cite{chen2025knowledge}, leverage local features to achieve high scalability in high-density wireless networks. However, as noted in recent studies on local routing \cite{ashvinkumar2022local}, these heuristics often fail in sparse networks where topological holes require detouring capabilities that exceed the local receptive field. Conversely, structural approaches typically employ Graph Neural Networks (GNNs) or spectral methods to capture global connectivity. For instance, RouteNet \cite{rusek2020routenet} demonstrates that Message Passing Neural Networks (MPNNs) can accurately model complex end-to-end delays and routing metrics. Similarly, spectral approaches utilizing Laplacian-based features \cite{steinerberger2021spectral}, specifically the resistance distance \cite{tetali1991random, liao2025efficient}, reliably identify shortest paths by mathematically encoding global network connectivity. However, these global methods often incur prohibitive computational and communication overheads for real-time routing. Our work seeks to reconcile this dichotomy by learning a policy that dynamically switches between these computationally distinct paradigms.

\subsection{Online Meta-Learning and Continual Adaptation}
Online meta-learning provides a robust framework for models to continuously adapt to non-stationary environments through incremental knowledge accumulation \cite{finn2017model}. Unlike traditional batch learning, online meta-learning strategies allow for rapid adaptation to new tasks as they arrive sequentially, which is critical for evolving network topologies. 
Recent developments in online cascade learning \cite{nie2024online} have further demonstrated that models can optimize performance over data streams by continuously updating their deferral policies. In the context of routing, Ca-MoE leverages this modularity to restrict fine-tuning to only the most relevant experts. This selective updating acts as a stability-focused mechanism, mitigating the catastrophic forgetting often observed in continual learning scenarios \cite{kirkpatrick2017overcoming}, while preserving the model's specialized knowledge across diverse graph environments.

\section{Problem Statement}
We consider a class of graphs $\mathbb{G}$ where each graph $G=(V, E)$ consists of $n$ nodes distributed according to a uniform random distribution in a 2-dimensional Euclidean square region. Each node $v \in V$ is locally aware of its global coordinates. An edge $(v,u) \in E$ exists if and only if the Euclidean distance between them does not exceed a predefined communication radius $R$. The network density is denoted by $\rho$, defined as the average number of nodes within an area of $R^2$. Consequently, the side length of the square region is given by $\sqrt{n R^2 / \rho}$.

\subsection{All-Pairs Near-Shortest Path Problem (APNSP)}
The objective of the APNSP problem is to derive a routing policy $\pi$ for any graph $G \in \mathbb{G}$ that identifies paths between all node pairs $(O, D)$ that are close to the optimal shortest path. A path is considered ``near-shortest" if its total length falls within a specified approximation factor, $\epsilon \ge 0$, of the actual shortest path length.

Let $d_e(O, D)$ represent the Euclidean distance and $d_{sp}(O, D)$ represent the shortest path length between an origin $O$ and destination $D$. We define the path stretch $\zeta(O, D)$ as the ratio:
\begin{equation}
    \zeta(O, D) = \frac{d_{sp}(O, D)}{d_e(O, D)}
\label{path_stretch}
\end{equation}

\textbf{The APNSP Problem Definition:}
We seek to learn a routing policy $\pi(O, D, v) = u$ that selects the next hop $u$ to construct a path $p(O,D)$. The goal is to maximize the routing accuracy, defined as the proportion of paths that satisfy the near-shortest constraint:
\begin{align}
    \max ~~ \text{Accuracy}_{G, \pi} &= \frac{1}{|V|^2} \sum_{O, D \in V} \eta(O, D),
\end{align}
where the indicator function $\eta(O, D)$ is defined as:
\begin{align}
    \begin{cases} 
    1, & \text{if } \frac{d_p(O, D)}{d_{sp}(O, D)} \leq \zeta(O, D)(1 + \epsilon) \\
    0, & \text{otherwise}
    \end{cases}.
\label{APNSP_accuracy_condition}
\end{align}

The inclusion of the stretch factor $\zeta(O, D)$ serves to normalize performance across varying network densities, which is critical in sparse networks where topological holes cause significant variance in path lengths.

\subsection{MDP Formulation for the APNSP Problem}
To solve the APNSP problem, we model the routing process as a Markov Decision Process (MDP). We assume a deterministic state transition $P(s'|s,a)=1$, where a state $s$ represents the features of the current node $v$ holding a packet for destination $D$, and an action $a \in A(s)$ represents the selection of a neighbor $u$ as the next forwarder.


To align the MDP objective with the shortest path goal, we define the instantaneous reward $r(s, a)$ as the negative Euclidean length of the edge $(v, u)$. We set the discount factor $\gamma = 1$. The $Q$-value, representing the total expected reward from state $s$ following action $a$, is defined as:
\begin{equation}
    Q(s_t, a_t) = \sum_{i=t}^{L} r(s_i, a_i) = - \sum_{i=t}^{L} d_e(v_i, v_{i+1}),
\label{Q_equation}
\end{equation}
where $L$ is the number of hops to the destination. Consequently, the optimal $Q$-value, $Q^*(s, a)$, corresponds to the negative length of the shortest path from the current node to the destination.

We employ Deep Neural Networks (DNNs) to approximate these optimal $Q$-values via a data-driven approach. By selecting actions that yield the highest $Q$-values (i.e., the least negative cumulative distance), the policy minimizes the total path length, thereby satisfying the APNSP accuracy condition defined in Eq.~\ref{APNSP_accuracy_condition}.

\section{Ca-MoE Framework for Solving APNSP}

\begin{figure*}[thb!]
    \centering
     \subfigure[$v$ in sparse graphs]
     {
        \includegraphics[width=0.3\textwidth]{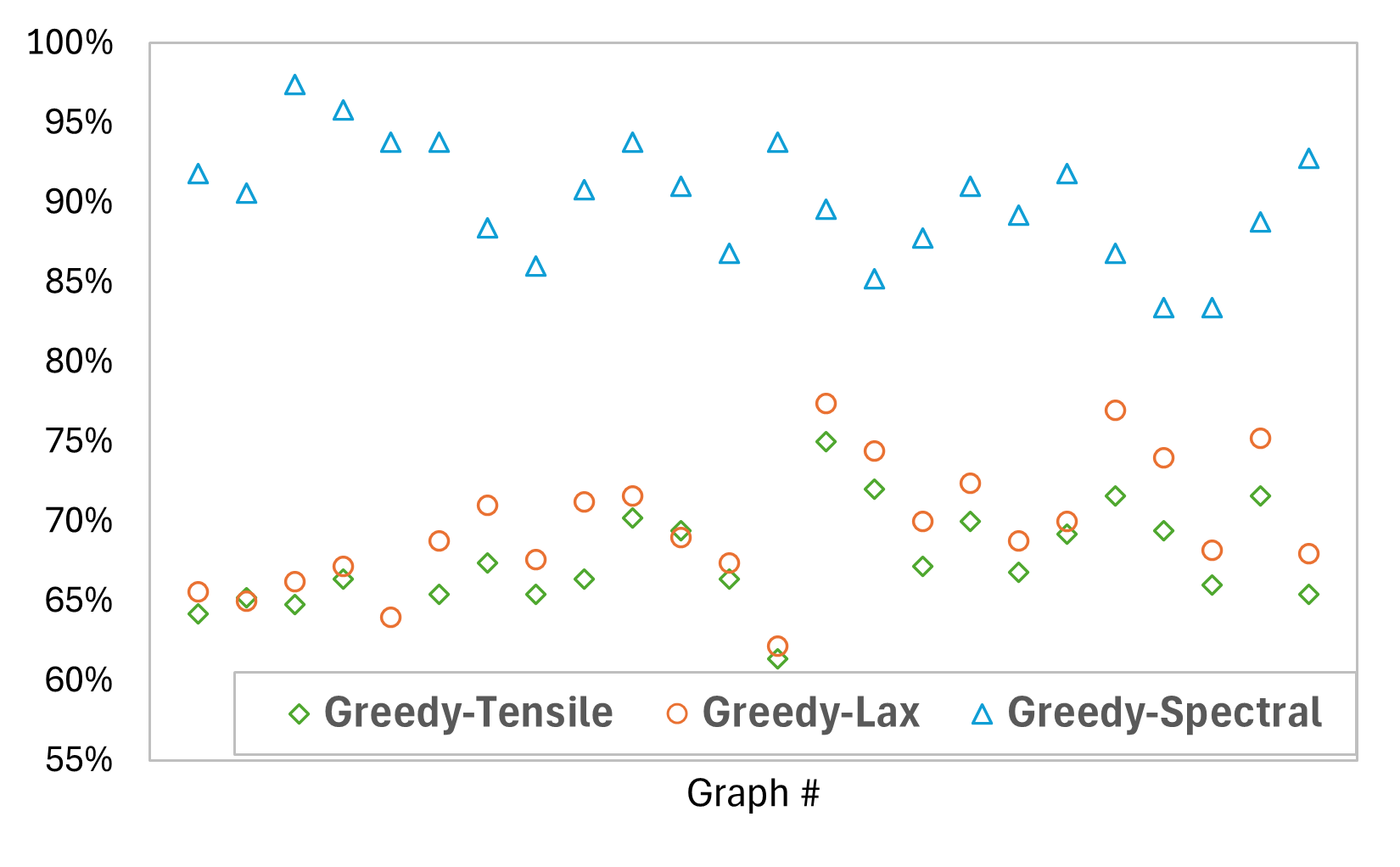}
        \label{Q_value_comparison_sparse_graphs_sampling_random_nodes}
    }
    \subfigure[$v$ near shortest paths in dense graphs]
     {
        \includegraphics[width=0.3\textwidth]{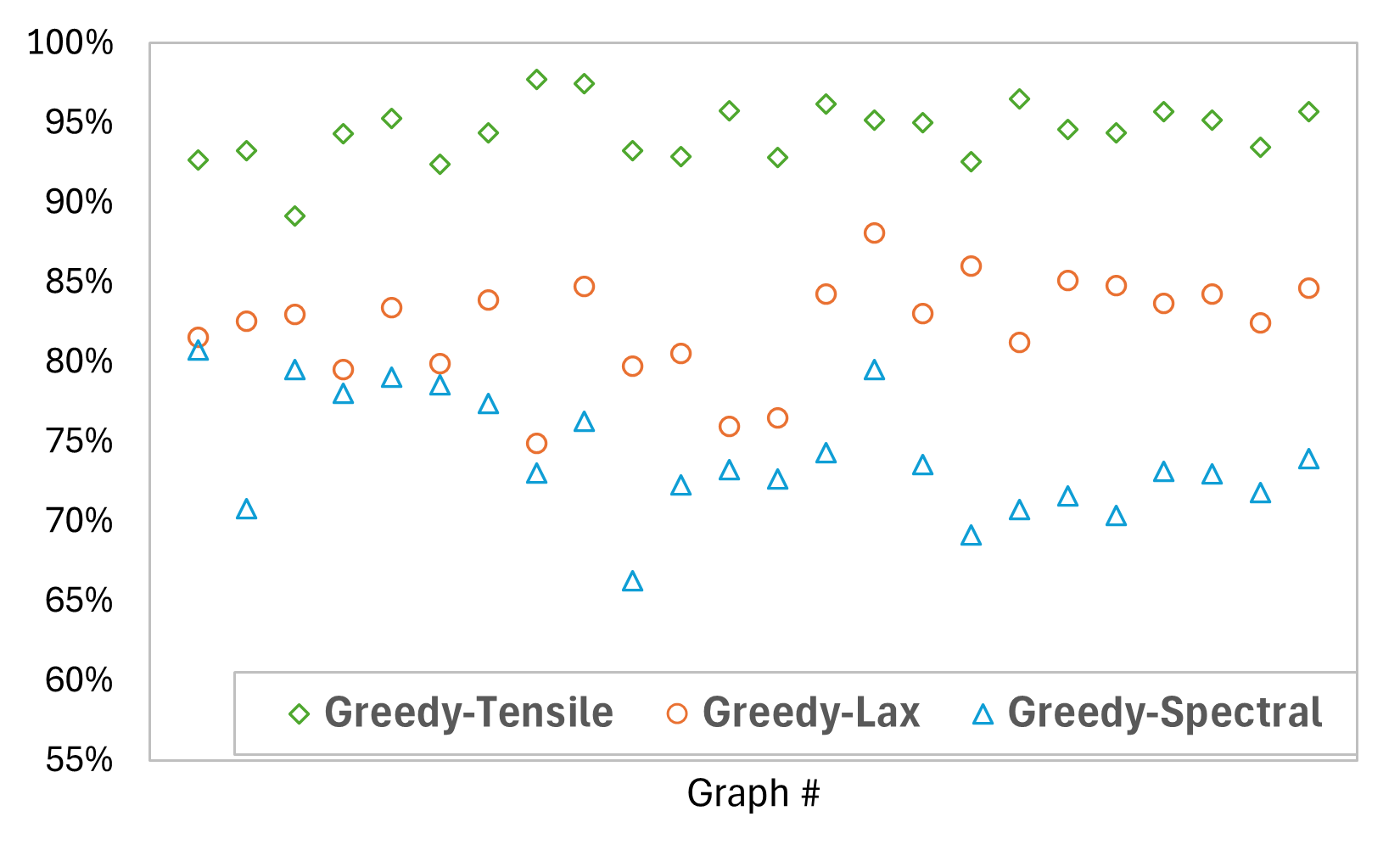}
        \label{Q_value_comparison_dense_graphs_sampling_nodes_on_SPs}
    }
    \subfigure[$v$ far from shortest paths in dense graphs]
     {
        \includegraphics[width=0.3\textwidth]{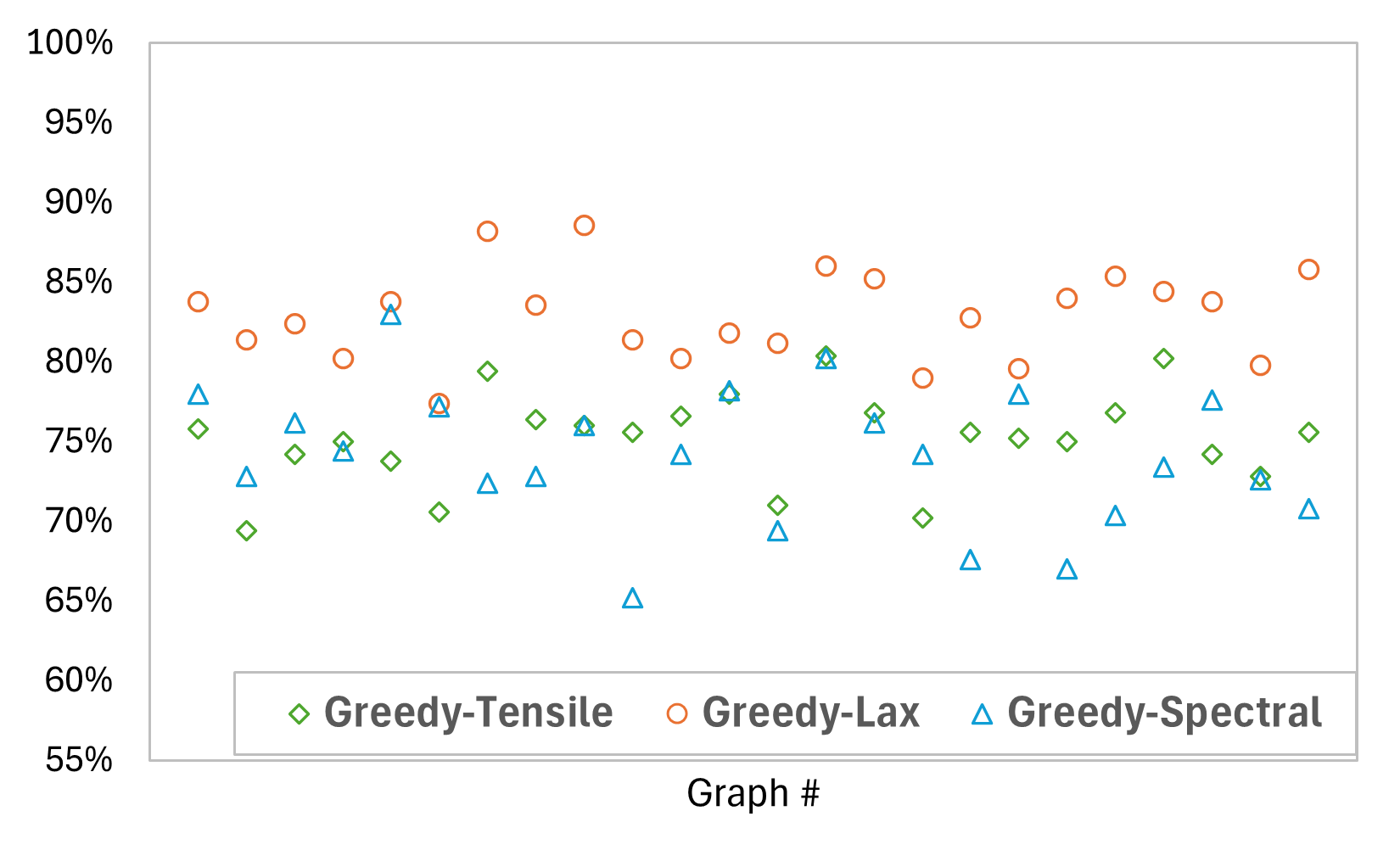}
        \label{Q_value_comparison_dense_graphs_sampling_nodes_not_on_SPs}
    }
    \caption{Comparative performance of the {\em Greedy-Tensile}, {\em Greedy-Lax}, and {\em Greedy-Spectral} routing policies across different densities of Euclidean graphs. Each sub-figure illustrates a set of graphs where a different routing policy dominates: different graphs are enumerated on the x-axis; for each graph, the y-axis denotes the frequency with which each expert selects the best next forwarder for nodes $v$ given samples $\langle O,D,v,nbr(v) \rangle$ from that graph.  In sub-figures (b) and (c), the choices for $v$ are constrained relative to the shortest paths from origin $O$ to $D$. Recall that a forwarder is determined to be the  ``best'' if it yields the maximal optimal $Q$-value, $Q^*(O, D, v, u)$.}
    \label{Q_value_comparison}
\end{figure*}

\subsection{Overview}


As shown in Figure~\ref{Ca-MoE}, the Cascaded Mixture of Experts (Ca-MoE) framework employs a modular two-tier architecture designed to balance computational efficiency with the generalization required for the all-pairs near-shortest path (APNSP) problem. By decoupling decision-making, Ca-MoE allows routine routing decisions to be handled by lightweight, lower-tier experts using local features, while computationally intensive upper-tier experts are adaptively triggered only when topological irregularities necessitate global structural awareness.

The forwarder selection process employs a cascading strategy. The {\bf Lower-Tier} comprises multiple lightweight experts that infer ranking metrics ($Q$-values) using local Euclidean features. A {\bf Router} dynamically assigns the input to the most suitable expert via a Top-1 strategy. To ensure reliability, a {\bf Deferral Rule} evaluates the quality of this initial decision. If the predicted next-hop forwarder is deemed insufficient ---typical of sparse regions with structural holes--- the task is cascaded to the {\bf Upper-Tier}. This tier activates a resource-intensive expert that integrates global spectral features to refine the decision\footnote{To maintain parameter efficiency, Ca-MoE employs a single upper-tier expert, as empirical evidence suggests that subsequent tiers yield diminishing returns on routing accuracy.}.

To support continuous adaptation, Ca-MoE employs an online meta-learning strategy that incrementally fine-tunes experts as new graph environments are encountered. This modularity enables independent updates using small sample sets, thereby minimizing retraining costs by isolating adjustments to the relevant experts. To ensure stability, the framework maintains an evaluation dataset ($ED$), which is dynamically updated to verify that fine-tuned models retain high performance on previously seen instances. By conditioning expert updates on this validation, Ca-MoE progressively accumulates meta-knowledge, yielding stronger generalization across diverse graph classes without the overhead of full model retraining.

\begin{figure*}[thb!]
    \centering
     \subfigure
     {       \includegraphics[width=0.89\textwidth]{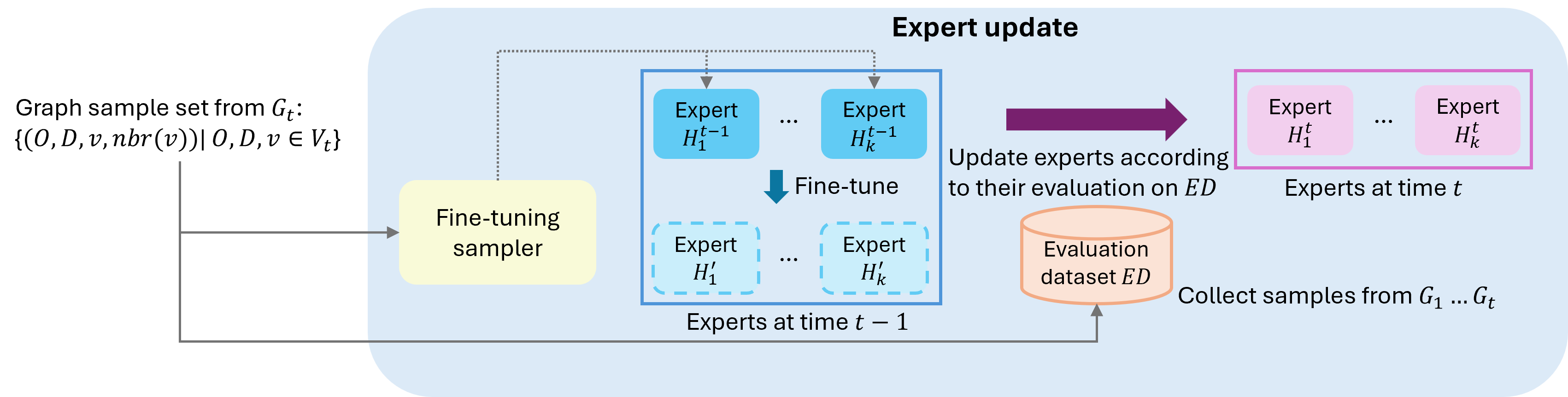}
    }
    \caption{The online meta-learning pipeline. Upon the arrival of a new graph $G_t$, experts are selectively fine-tuned via a fine-tuning sampler. To prevent catastrophic forgetting, updated models ($H'$) are validated against a persistent Evaluation Dataset ($ED$) before replacing the current experts ($H^{t-1}$).}
    \label{online_meta_learning}
\end{figure*}

\subsection{Lower-Tier Experts with Local Knowledge}
The Lower-Tier comprises two distinct experts ---{\em Greedy-Tensile} ($H_1$) and {\em Greedy-Lax} ($H_2$)--- and a router that dynamically assigns the most suitable expert for a given graph sample $\langle O, D, v, nbr(v) \rangle$.



\subsubsection{Complementary Expert Strategies} 
The deployment of multiple experts within the lower-tier is non-trivial and is necessitated by the distinct forwarding behaviors required in differing local contexts, as empirically demonstrated in Figures 2(b) and 2(c). Because a lightweight policy is not readily optimized for both routine traversal and obstacle avoidance, we architect two complementary experts to handle specific scenarios of routing:

\begin{itemize}
\item {\em Greedy-Tensile} ($H_1$)---Routine Forwarding: Optimized for standard Euclidean environments, this expert prioritizes low-stretch forwarding. It serves as the primary policy when the current node $v$ lies proximal to the shortest path between $O$ and $D$ (i.e., the ``on-track" scenario). 

\item {\em Greedy-Lax} ($H_2$)---Recovery Forwarding: Designed to address the limitations of {\em Greedy-Tensile}, this expert allows higher-stretch forwarding decisions. When a packet is forced to deviate from the shortest path due to topological holes, {\em Greedy-Lax} functions as a specialized recovery strategy. By allowing trajectories that temporarily diverge from the direct line to the destination, it effectively navigates around obstacles to steer the packet back toward a routine forwarding region.
\end{itemize}

\subsubsection{Feature Selectors for \GreedyTensile and Greedy-Lax}
To maintain computational efficiency, both common experts utilize a shared set of local features to construct their input vectors. For a routing request with origin $O$ and destination $D$, where the packet is currently at node $v$, the feature selector constructs an input vector based on the following embeddings:

\begin{itemize}
\item State feature, $f_{s}(v)$: A vector characterizing the current node $v$ relative to the destination:
    \begin{enumerate}[label=(\alph*)]
    \item \textbf{Distance to destination: $d(v, D)$}, the distance between $v$ and $D$.
    \item \textbf{Node Stretch: $ns(O, D, v) = \frac{d(O, v)+d(v, D)}{d(O, D)}$}. This ratio represents the deviation of the path through $v$ relative to the direct Euclidean distance between $O$ and $D$.. 
    \end{enumerate}

\item Action feature, $f_a(u)$: For each candidate neighbor $u \in nbr(v)$, the action feature corresponds to the state feature of that neighbor, such that $f_a(u) = f_s(u)$.
\end{itemize}

The inputs to the expert networks are the concatenated vectors $\langle f_s(v), f_a(u) \rangle$, which are used to predict the $Q$-value for forwarding to neighbor $u$.

\subsubsection{Training of Router}\label{router_training}
The router employs a Top-1 strategy to select the expert that maximizes the expected routing efficiency (i.e., the one predicting the highest valid $Q$-value). Unlike the experts, which focus on evaluating specific neighbors, the router evaluates the broader context of the node to determine which policy ({\em Greedy-Tensile} or {\em Greedy-Lax}) is required.

Upon the arrival of a new graph $G_t$, the router undergoes incremental retraining using a sample set:
\\ 
\hspace*{10mm}
$X \! = \! \bigcup_{v \in V_{sample}}{\{feat_{router}(O,D,v,nbr(v))\}}$ and 
\\ 
\hspace*{10mm}
$Y \! = \!  \bigcup_{v \in V_{sample}}{y_{v} \in \{0,1\} }$, \\
where $y_v$ is a binary label indicating the expert that yielded the superior routing decision for that instance. The router's feature vector, $feat_{router}$, encapsulates broader local context, including:
\begin{itemize}
\item Network density ($\rho$) and the node degrees of $v$.
\item Node stretch ($ns(v, O, D)$) of the current node $v$ relative to the $(O, D)$ pair..
\item The Euclidean distances $d(O, D)$, $d(O, v)$, and $d(v, D)$.
\end{itemize}

This multi-dimensional input allows the router to learn a meta-policy ---for example, preferring {\em Greedy-Lax} in lower-density regions where detours are likely required--- thereby minimizing the computational overhead of the inference process.

\subsection{Upper-Tier Expert with Global Knowledge}
While the lower-tier handles the majority of routine routing decisions, certain graph instances ---particularly those in sparse regions or containing significant topological irregularities--- require broader structural awareness that local features cannot provide. The upper-tier is designed to resolve these complex scenarios by adaptively activating the \GreedySpectral ($H_3$) expert.

\subsubsection{Global Expert Strategy ($H_3$)}
The necessity of the upper-tier is empirically substantiated by Figure~\ref{Q_value_comparison_sparse_graphs_sampling_random_nodes}, where local-feature experts (\GreedyTensile and {\em Greedy-Lax}) exhibit significant performance degradation in sparse environments. In these scenarios, topological holes frequently trap local heuristics in local minima. To resolve these bottlenecks, the upper-tier activates the \GreedySpectral expert.

\GreedySpectral incorporates global structural knowledge by leveraging the resistance distance $\Omega(i, j)$. Derived from the graph Laplacian matrix, this metric mathematically encodes the global connectivity between nodes. By optimizing for resistance distance, \GreedySpectral effectively identifies forwarders that avoid dead-end or excessively long trajectories, ensuring structurally reliable forwarding in complex topologies where lower-tier experts may fail.

\subsubsection{Feature Selector for Greedy-Spectral}
To enable this global reasoning, the feature selector constructs input vectors based on resistance distance. For a routing request $(O, D)$ at current node $v$:

\begin{itemize}
    \item State feature: $$f_s(v) = \Omega(v, D)$$.
    \item Action feature: $$f_a(u) = \Omega(u, D)$$ for each neighbor $u \in nbr(v)$.
\end{itemize}


\subsubsection{Computational Feasibility}
Although calculating the full resistance distance matrix scales cubically ($O(N^3)$), the proposed framework requires only specific pairwise queries. In sparse networks, the utilization of nearly-linear time Laplacian solvers \cite{spielman2014nearly} and labeling schemes optimized for small-treewidth graphs \cite{liao2025efficient} reduces the query complexity to a near-linear complexity of $\tilde{O}(|E|)$. This efficiency ensures that global feature extraction remains feasible for real-time routing when the upper-tier is activated.



\subsection{Adaptive Activation via Deferral Rule}
The Deferral Rule serves as a lightweight gating mechanism that governs the transition between the local and global tiers. Implemented as a pre-trained binary classification DNN, it evaluates whether the forwarder $\hat{u}$ selected by the lower-tier expert is likely to yield a globally near-shortest path.

During online inference, the model relies exclusively on the local observation vectors $X$ (as defined in Section~\ref{router_training}) to approximate global path quality. This design bridges the information gap, allowing the system to predict global optimality without incurring the cost of global state maintenance.

To bridge this gap, the deferral rule is trained using ground-truth labels derived from global metrics. A positive label $y_v=1$ is assigned if the path through neighbor $u$ satisfies the near-optimality condition: $$d_{sp}(O, v) + d_{e}(v, u) + d_{sp}(u, D) \leq d_{sp}(O, D) \times (1+\Delta),$$ where $\Delta$ represents the tolerable deviation threshold. 


\subsection{Online Meta-learning}
To ensure the Ca-MoE framework adapts to evolving graph topologies without suffering from catastrophic forgetting, we employ an online meta-learning strategy. This process maintains a persistent Evaluation Dataset (ED) for stability validation and performs targeted fine-tuning on arriving graph instances to continuously update the experts.



\subsubsection{Evaluation Dataset ($ED$) Maintenance}
$ED$ serves as a consistency benchmark, preserving a diverse history of routing scenarios. Upon the arrival of a new graph $G_t$, we update the $ED$ to reflect the shifting distribution of network environments. We sample a small subset of instances $(O, D, v)$ from $G_t$ and compute the corresponding ground-truth $Q$-values, $Q^*(O, D, v, u)$, for all neighbors $u \in nbr(v)$. These samples are stored as tuples of state features, action features, and optimal $Q$-values: $\langle f_{s}(v), \{f_{a}(u)\}_{u}, \{Q^*\}_{u} \rangle$.  To prevent unbounded growth and ensure the dataset remains representative of both historical and recent samples, $ED$ operates as a fixed-size sliding window buffer.

\subsubsection{Targeted Expert Fine-Tuning}
To facilitate rapid adaptation, we derive expert-specific training sets from $G_{t}$ using differentiated sampling strategies. This ensures that each expert is fine-tuned only on scenarios relevant to its specialization, and prevents noise injection when relevant samples are unavailable:

\begin{itemize}
    \item \GreedyTensile $(H_1)$: This expert is fine-tuned on $(O, D)$ pairs characterized by low path stretch and short Euclidean distances. The training set minimizes prediction error in ``on-track" scenarios where local greedy heuristics suffice.
    
    \item \GreedyLax $(H_2)$ and \GreedySpectral $(H_3)$: These experts specialize in navigating topological irregularities. Consequently, their fine-tuning set consists of $(O, D)$ pairs exhibiting high path stretch and large source-destination distances. This exposes the models to non-routine cases (e.g., structural holes) where standard greedy forwarding fails.
\end{itemize}
If a graph $G_{t}$ does not contain samples matching an expert's criteria (e.g., a dense graph yielding no high-stretch shortest paths), fine-tuning for that specific expert is skipped.



\subsubsection{Stability-Aware Update Mechanism}
We employ a rigorous validation step to determine if updated parameters should be committed. Let $H_{i}^{t-1}$ denote an expert at time $t-1$ and $H_{i}^{\prime}$ denote the candidate fine-tuned version. Both models are evaluated on the $ED$ using the Discounted Cumulative Gain (DCG) metric, which measures the alignment between the predicted ranking of neighbor nodes and the ground-truth ranking derived from $Q^{*}$. The update is governed by a strict improvement criterion:
\begin{align*}
    H_{i}^{t} =  \begin{cases} H_{i}^{'} ,  \ \ \text{if} \ \text{Eval}(H_{i}^{'}, ED) > \text{Eval}(H_{i}^{t-1}, ED)\\
    H_{i}^{t-1},  \ \ \text{otherwise},
    \end{cases}.
\end{align*}
where $Eval(\cdot, ED)$ calculates the average ranking similarity across all sampled instances in the $ED$. This conditional mechanism ensures that the meta-model evolves only when the candidate expert demonstrates superior ranking fidelity across both historical and current data




\begin{table}[htbp]
\caption{Simulation Parameters}
\begin{center}
\begin{tabular}{c c c}
\hline
\textbf{Symbol}& \textbf{Meaning} & \textbf{Value} \\
\hline
$N_{train}$ & size of seed graphs for training experts & 50 \\
\hline
$\rho_{GT}$ & density of seed graph & 5 \\
 & for training \GreedyTensile & \\
\hline
$\rho_{GL/GS}$ & density of seed graphs  & 2 \\ 
& for training \GreedyLax and \GreedySpectral & \\
\hline
$R$ & communication radius & 1000 \\
\hline
$I_{GT/GL}$ & \# of input features  & 4 \\
& for \GreedyTensile and \GreedyLax & \\
\hline
$I_{GS}$ & \# of input features  & 2 \\
& for \GreedySpectral & \\
\hline
$K$ & \# of hidden layers for DNNs & 2 \\
\hline
$N_{e}[]$ & \# of neurons in each hidden layer  & $[50I, I]$ \\
\hline
$\epsilon$ & margin for shortest paths prediction & 0.05 \\
\hline
$\phi$ & \# of sampled nodes for updating $ED$ & 10\\
\hline
$IterNum_{FT}$ & \# of iterations in fine-tuning & 1000 \\ 
\hline
\end{tabular}
\label{SimulationParams}
\end{center}
\end{table}

\begin{figure*}[thb!]
    \centering
     \subfigure[Graphs where \GreedyTensile underperforms]
     {
        \includegraphics[width=0.31\textwidth]{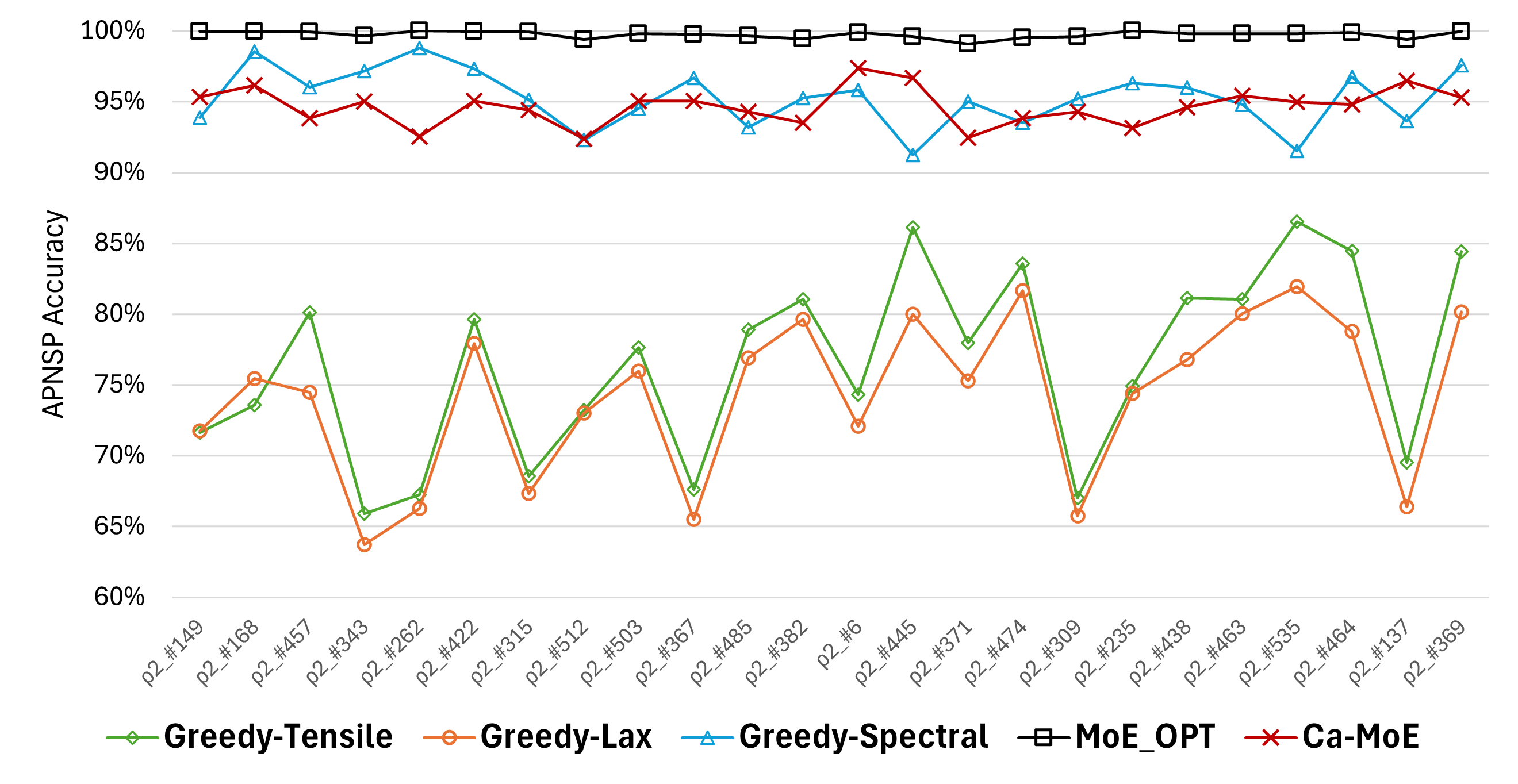}
        \label{Eval_APNSP_accuracy_on_single_graphs_low_sim_graphs}
    }
    \subfigure[Graphs where \GreedyTensile outperforms]
     {
        \includegraphics[width=0.31\textwidth]{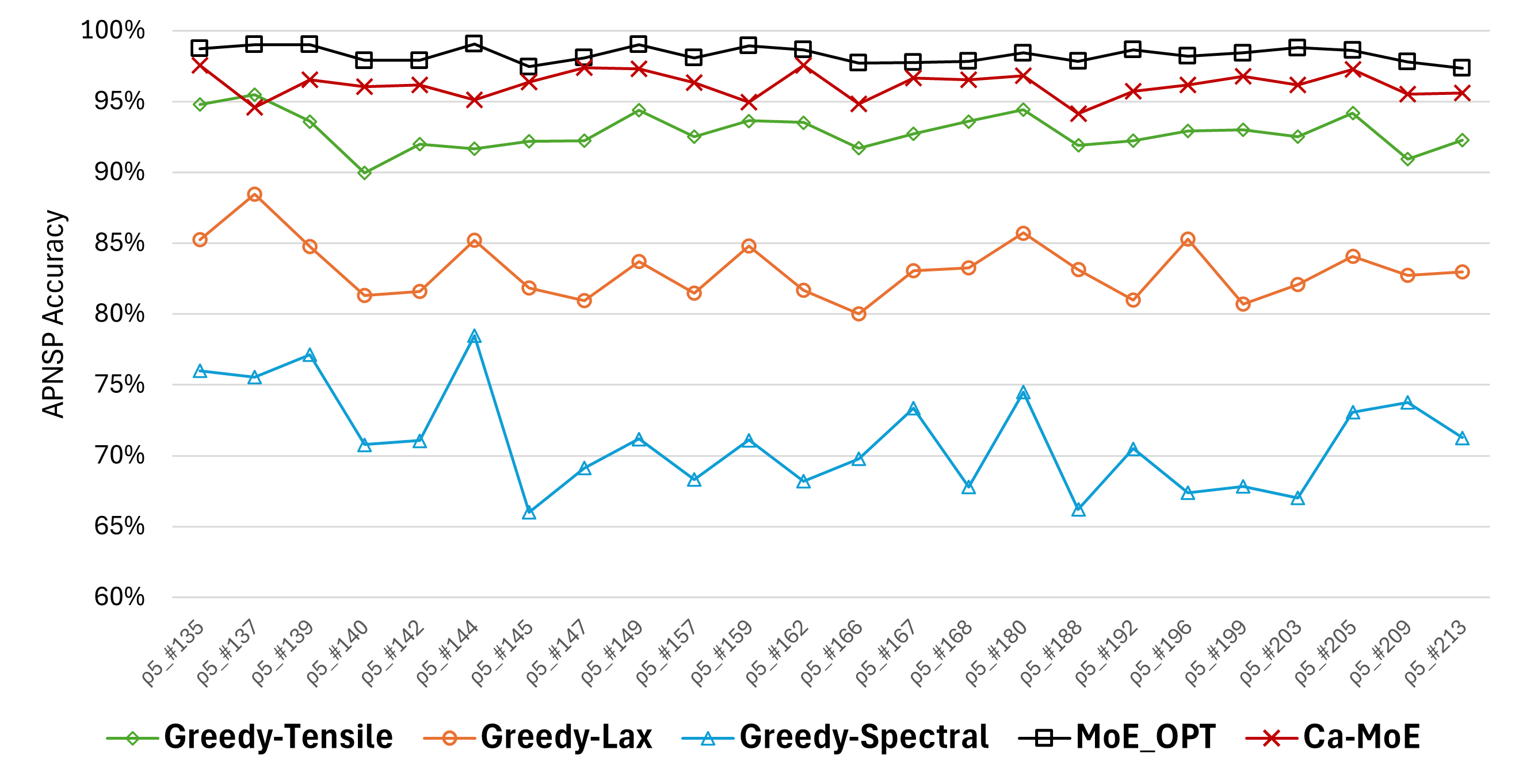}
        \label{Eval_APNSP_accuracy_on_single_graphs_dense_graphs}
    }
    \subfigure[Graphs with various density $\rho \in \{2, 5\}$]
     {
        \includegraphics[width=0.31\textwidth]{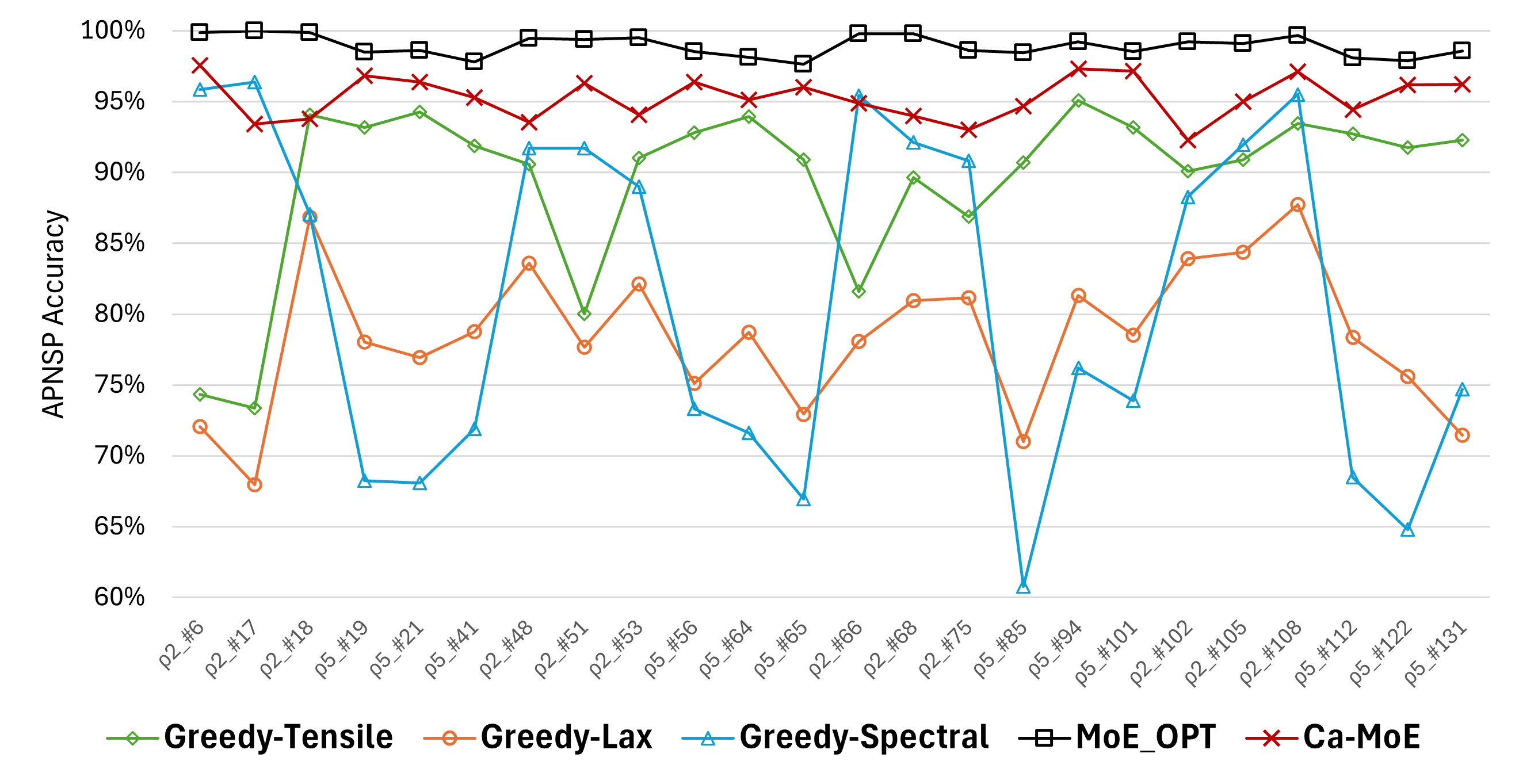}
        \label{Eval_APNSP_accuracy_on_single_graphs_mixed_graphs}
    }
    \caption{Comparative APNSP prediction accuracy on individual graph instances ($G_t$) for Ca-MoE against single-expert baselines. The x-axis enumerates specific graph instances, while the y-axis reports the routing accuracy.}
    \label{Eval_APNSP_accuracy_on_single_graphs}
\end{figure*}

\begin{figure*}[thb!]
    \centering
     \subfigure[Graphs where \GreedyTensile underperforms]
     {
        \includegraphics[width=0.31\textwidth]{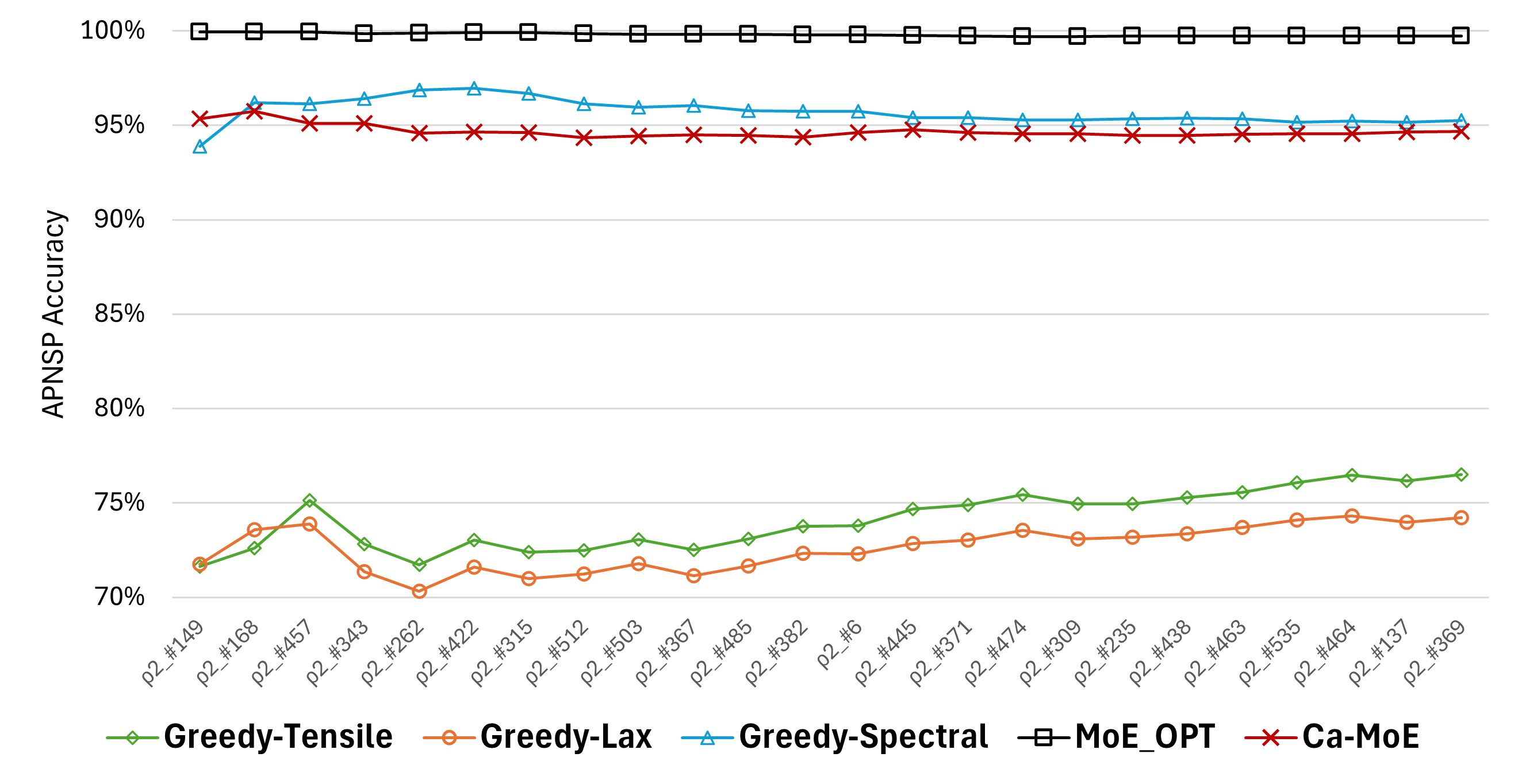}
        \label{Eval_APNSP_accuracy_on_seen_graphs_low_sim_graphs}
    }
    \subfigure[Graphs where \GreedyTensile outperforms]
     {
        \includegraphics[width=0.31\textwidth]{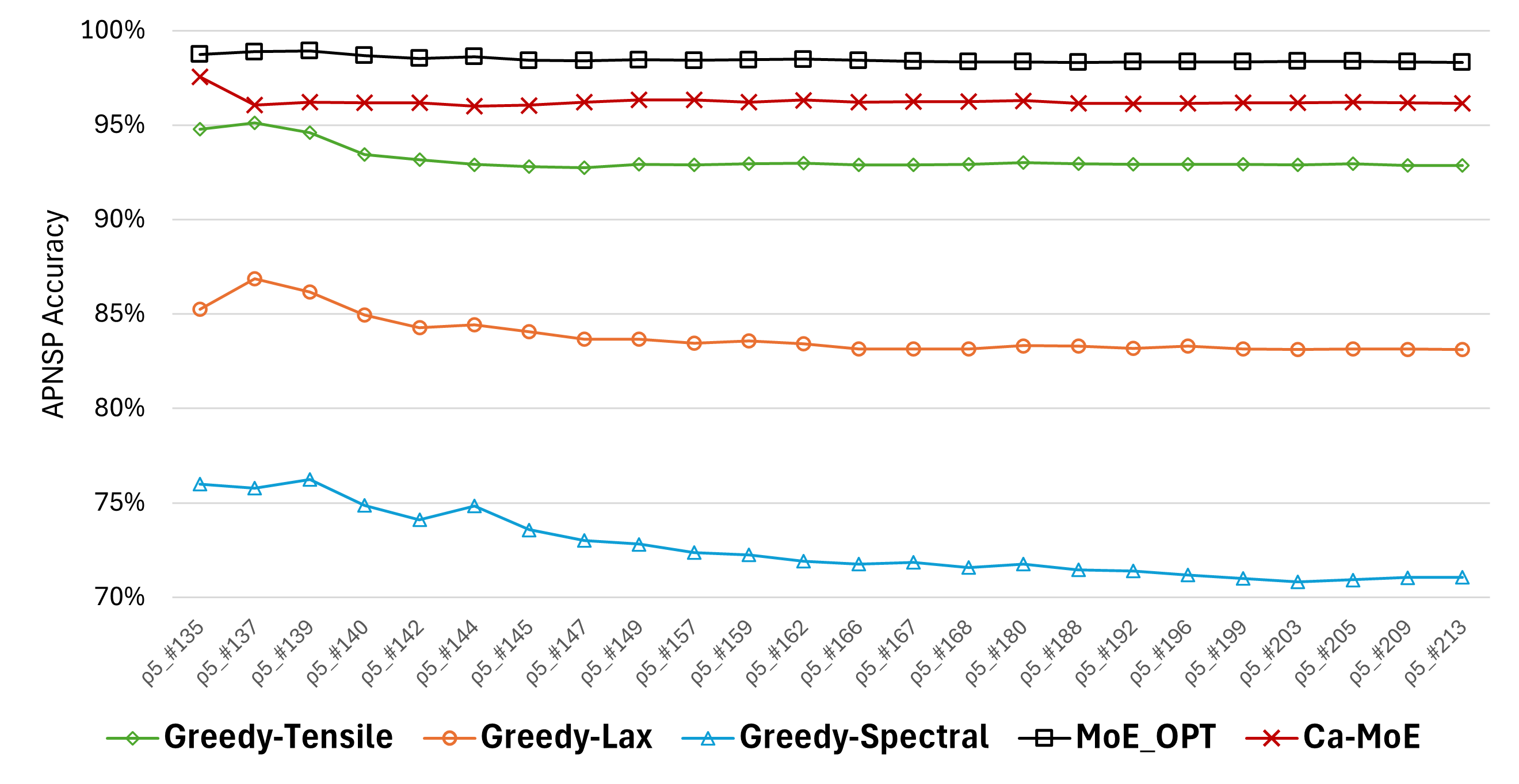}
        \label{Eval_APNSP_accuracy_on_seen_graphs_dense_graphs}
    }
    \subfigure[Graphs with various density $\rho \in \{2, 5\}$]
     {
        \includegraphics[width=0.31\textwidth]{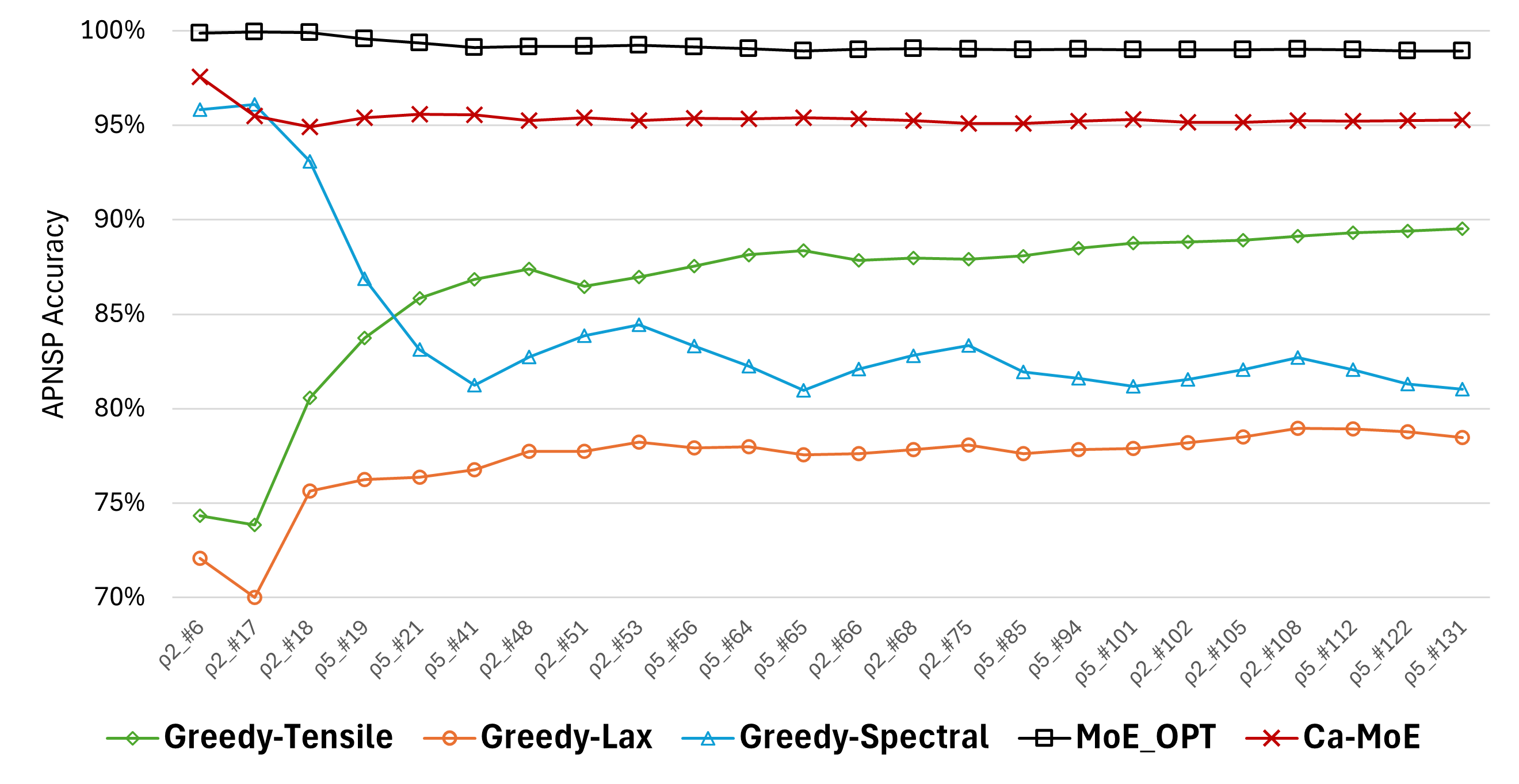}
        \label{Eval_APNSP_accuracy_on_seen_graphs_mixed_graphs}
    }
    \caption{Evolution of average APNSP prediction accuracy across the sequence of seen graphs $G_1..G_t$ shown in Figure~\ref{Eval_APNSP_accuracy_on_single_graphs}. The figure illustrates the continual generalization capability of the framework. The x-axis represents the chronological sequence of arriving graphs, and the y-axis denotes the average accuracy on all graphs encountered up to time $t$.}
    \label{Eval_APNSP_accuracy_on_seen_graphs}
\end{figure*}

\section{Performance Evaluation}
In this section, we evaluate the Ca-MoE framework in predicting all-pairs near-shortest paths across heterogeneous graph topologies. We assess its ability to balance accuracy with generalization compared to single-expert baselines and a theoretical upper bound.

{\bf Implementation Details:} All components, including the experts and the router, are implemented as deep neural networks (DNNs) using PyTorch 2.9.0. The simulation parameters for initial training and online fine-tuning are detailed in Table~\ref{SimulationParams}.

\subsection{Comparative Baselines}
To assess the efficacy of the cascading architecture, we compare Ca-MoE against the following policies:

\begin{itemize}
\item {\em Greedy-Tensile~\cite{chen2025knowledge}}: A local-feature baseline optimized for on-track scenarios with low path stretch.

\item {\em Greedy-Lax}: A secondary local expert designed for high-stretch recovery forwarding when nodes deviate from the shortest path.

\item {\em Greedy-Spectral}: A global-knowledge expert using resistance distance features, designed for high accuracy in sparse or complex environments.

\item {\em MoE\_OPT}: An idealized oracle that perfectly selects the best-performing expert for every routing decision, serving as the theoretical upper bound for the modular framework.
\end{itemize}

\subsection{Generalization over Diverse Graphs}
We benchmark the framework across three distinct scenarios to test robustness against topological shifts:

\begin{enumerate}[label=(\alph*)]
\item Sparse/Challenging Topologies: Graphs where local heuristics ({\em Greedy-Tensile}) typically underperform (accuracy ${\tiny \sim} \! \! ~75\%$).

\item Dense Topologies: Graphs where local knowledge is generally sufficient (accuracy $> 90\%$).

\item Mixed Densities: A set spanning both sparse and dense configurations ($\rho \in \{2, 5\}$) to evaluate meta-knowledge accumulation.
\end{enumerate}


\subsubsection{Single-Graph Prediction Accuracy}
Figure~\ref{Eval_APNSP_accuracy_on_single_graphs} illustrates the routing accuracy on individual graph instances. In sparse networks containing topological holes (Figure~\ref{Eval_APNSP_accuracy_on_single_graphs_low_sim_graphs}), local-feature experts degrade significantly. In contrast, Ca-MoE achieves parity with the global {\em Greedy-Spectral} expert, improving accuracy by up to 29.1\% over {\em Greedy-Tensile}.

Even in dense topologies where local heuristics dominate (Figs.~\ref{Eval_APNSP_accuracy_on_seen_graphs_dense_graphs} and ~\ref{Eval_APNSP_accuracy_on_seen_graphs_mixed_graphs}), Ca-MoE yields a performance gain of up to 6\%. This indicates that the router successfully identifies specific nodes requiring specialized recovery strategies ({\em Greedy-Lax}) or global context ({\em Greedy-Spectral}), even within generally simple graphs. Notably, Ca-MoE consistently maintains performance within 1\%–6\% of the theoretical upper bound (MOE\_OPT).



\subsubsection{Continual Generalization}
Figure~\ref{Eval_APNSP_accuracy_on_seen_graphs} demonstrates the efficacy of the online meta-learning strategy. As new graphs are encountered, Ca-MoE progressively accumulates and maintains structural knowledge.

\vspace*{-1mm}
\begin{itemize}
    \item Sparse networks: The framework maintains $\sim \! 95\%$ generalization accuracy across previously seen sparse graphs (Figure~\ref{Eval_APNSP_accuracy_on_seen_graphs_low_sim_graphs}).

    \item Dense networks: It maintains $\sim \! 96\%$ accuracy across dense environments (Figure~\ref{Eval_APNSP_accuracy_on_seen_graphs_dense_graphs}).

    \item Robustness: Most importantly, it demonstrates high adaptability in mixed environments (Figure~\ref{Eval_APNSP_accuracy_on_seen_graphs_mixed_graphs}), maintaining $\sim \! 96\%$ accuracy even when the network density shifts drastically between $\rho=2$ and $\rho=5$.

    \item Larger networks: It maintains high accuracy for larger networks (we tested up to $n$=200 nodes).
\end{itemize}

\begin{figure}[thb!]
    \centering
     \subfigure
     {       \includegraphics[width=0.48\textwidth]{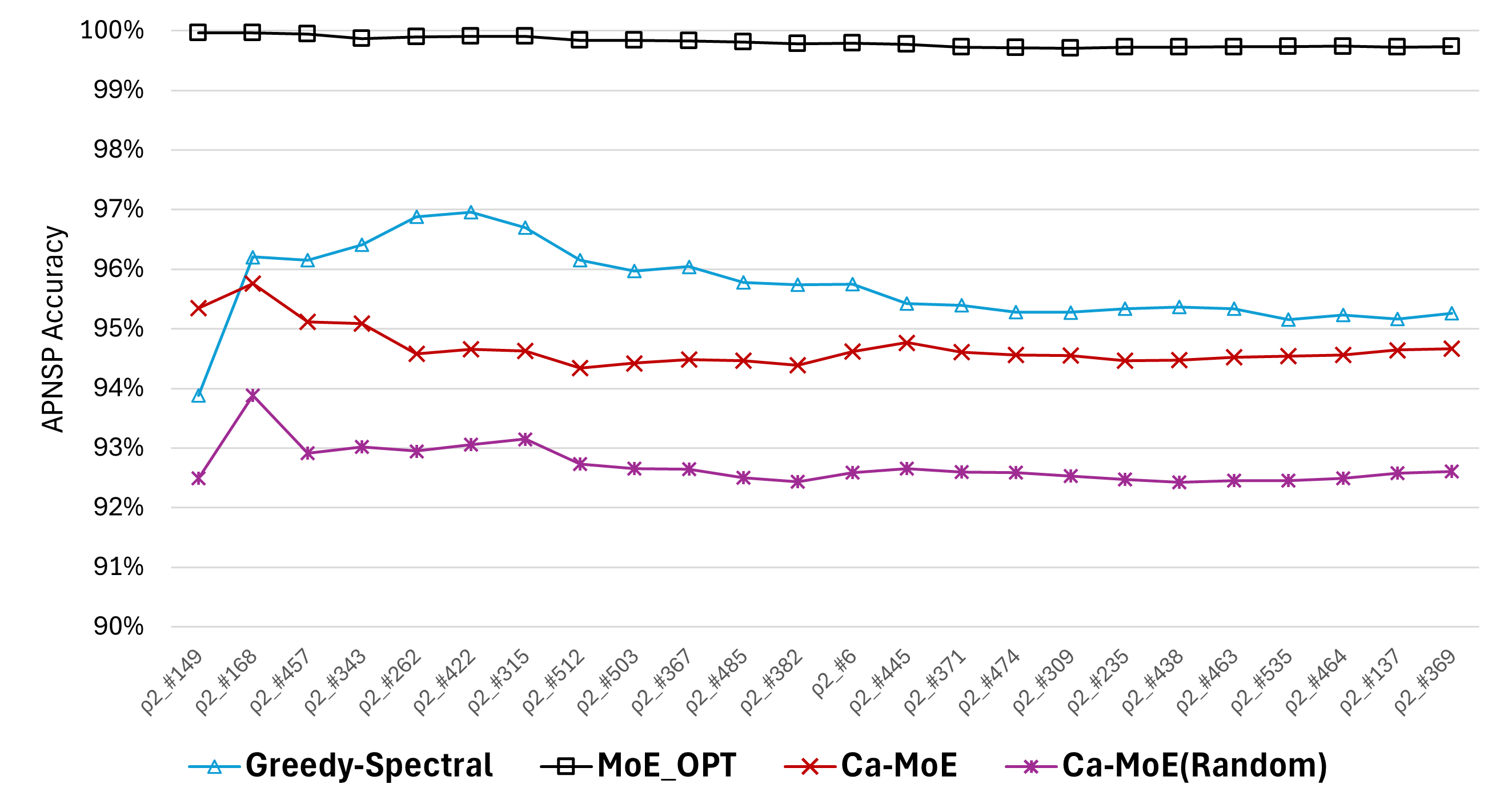}
    }
    \caption{Comparative APNSP prediction accuracy across the sequence of seen graphs $G_1..G_t$ shown in Figure~\ref{Eval_APNSP_accuracy_on_seen_graphs_low_sim_graphs}. This comparison highlights the framework's robustness to initialization by tracking the convergence of Ca-MoE(Random) (arbitrarily initialized experts) against the standard Ca-MoE framework and other baselines.}
    \label{Eval_APNSP_accuracy_on_seen_graphs_random_experts}
\end{figure}

\subsection{Robustness to Initialization}
To assess whether the Ca-MoE framework depends heavily on high-quality pre-training, we evaluated its convergence capabilities starting from a ``cold start" configuration. We introduced a specific variant, Ca-MoE (Random), where all expert models were initialized with arbitrary parameters rather than pre-trained weights. As illustrated in the experimental results (Figure ~\ref{Eval_APNSP_accuracy_on_seen_graphs_random_experts}), the performance of this randomly initialized model demonstrates remarkably rapid convergence. Despite beginning without prior knowledge, the model achieves an APNSP accuracy exceeding 90\% after observing only a small sequence of graph instances. This trajectory indicates that the online meta-learning strategy successfully extracts and accumulates meta-knowledge from incoming data streams, allowing the experts to progressively refine their policies. These findings confirm that the fine-tuning mechanism is highly data-efficient and effectively mitigates the cold-start problem, enabling the framework to adapt to new environments without the computational burden of extensive offline pre-training.

\subsection{MoE Overhead}

\begin{figure*}[thb!]
    \centering
     \subfigure[Graphs where \GreedyTensile underperforms]
     {
        \includegraphics[width=0.48\textwidth]{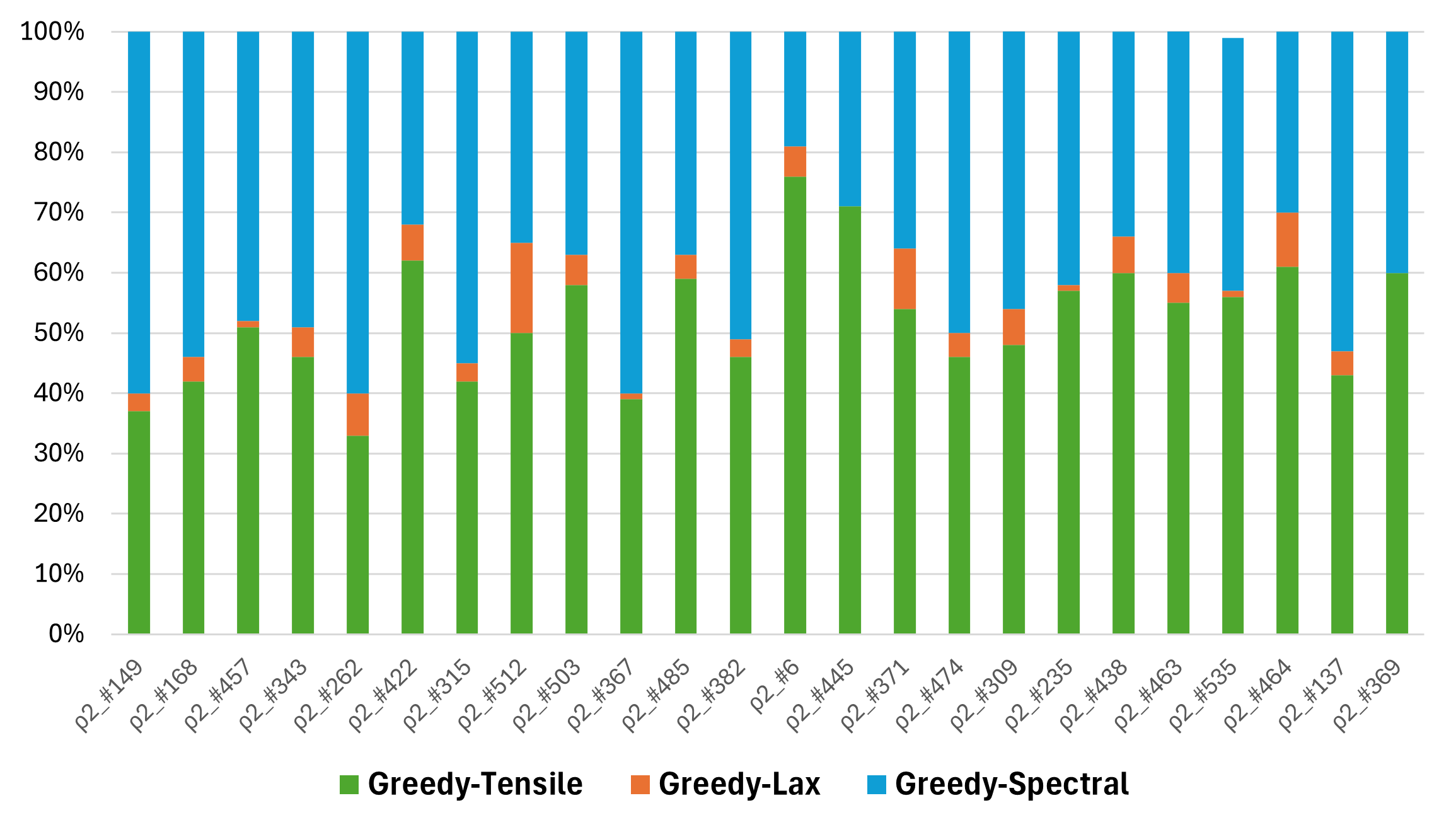}
        \label{Ca-MoE_expert_utilization_for_single_graph_low_SIM_graphs}
    }
    \subfigure[Graphs where \GreedyTensile outperforms]
     {
        \includegraphics[width=0.48\textwidth]{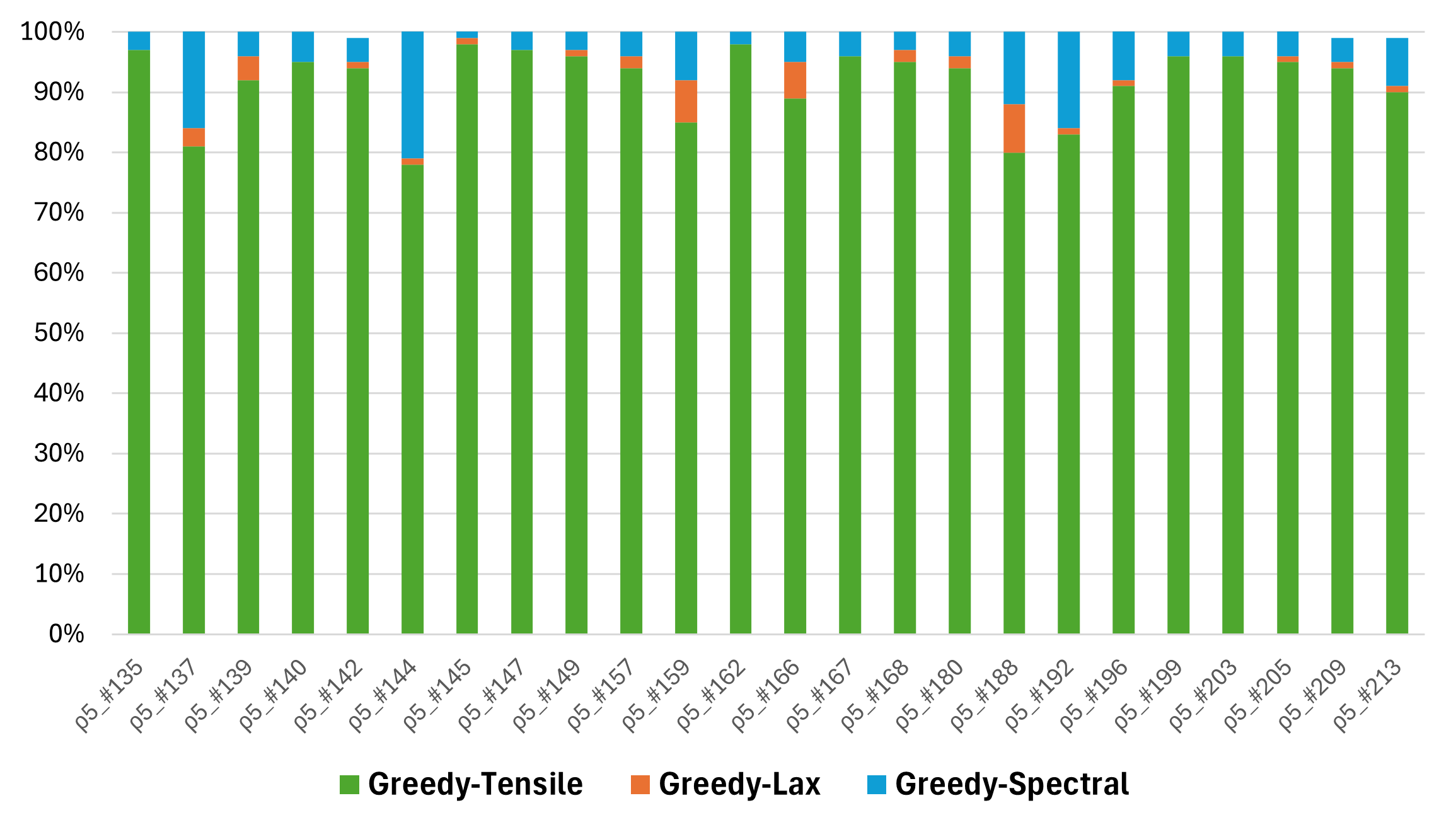}
        \label{Ca-MoE_expert_utilization_for_single_graph_dense_graphs}
    }
    \caption{Ca-MoE's expert utilization on individual graph instances. The x-axis enumerates specific graph instances, while the y-axis reports the the frequency with which each expert is activated for contributing the final selection of forwarder.}
    \label{Ca-MoE_expert_utilization_on_single_graphs}
\end{figure*}
A primary motivation for the Ca-MoE architecture is to minimize the computational overhead associated with global feature extraction by limiting its use to necessary scenarios. To quantify this efficiency, we analyzed the frequency with which each expert was selected to make the final forwarding decision across diverse topologies. (We note the overhead of the router and evaluator
is relatively small.) The utilization patterns shown in Figure~\ref{Ca-MoE_expert_utilization_on_single_graphs} reveal distinct behaviors based on network density.
Compared with the result in Figure~\ref{Ca-MoE_expert_utilization_for_single_graph_dense_graphs}, Figure~\ref{Ca-MoE_expert_utilization_for_single_graph_low_SIM_graphs} shows that the frequency of activating \GreedySpectral increases as the density decreases. However, compared to \GreedySpectral which has 100\% frequency to retrieve global feature, Ca-MoE avoids expensive computation in up to $\sim$80\% of the forwarding instances by only using \GreedyTensile or \GreedyLax for those graph samples. 
Notably, the utilization of \GreedyLax emerges as the demand for performing recovery forwarding in sparse network increases.
In dense networks, \GreedyTensile significantly dominate the utilization of key experts. This further implies the adaptive activation of Ca-MoE effectively prevent using resource-intensive expert in the network configurations where the necessity of using global knowledge for near-optimal routing is rare.



\section{Conclusion}
Experimental evaluations demonstrate that Ca-MoE significantly improves routing performance compared to single-expert baselines, achieving up to a 29.1\% accuracy gain in sparse graphs and consistently outperforming local-feature models in dense environments. Furthermore, our integrated online meta-learning strategy ensures that the framework continuously adapts to newly encountered graph classes without the overhead of full model retraining. 
We attribute the 1\%-6\% performance difference between {\em MoE\_OPT} and Ca-MOE to the intentional use of local knowledge only in training the router, for reasons of compute efficiency.
Individual experts can be made even more efficient by symbolic distillation. Overall, Ca-MoE provides a scalable and robust solution for near-shortest path routing, maintaining stability and high accuracy across diverse and evolving network distributions.  The framework is suitable for light-weight, near real time processing for dynamic networks, which are commonly found in mobile, ad hoc, and wireless networks.

\bibliographystyle{IEEEtran}
\bibliography{myref}

\end{document}